\documentclass[10pt,twocolumn,letterpaper]{article}

\usepackage{cvpr}              %

\usepackage[accsupp]{axessibility}  %

\usepackage{graphicx}
\usepackage{amsmath}
\usepackage{amssymb}
\usepackage{booktabs}
\usepackage{xcolor}
\usepackage{pifont}
\usepackage[utf8]{inputenc}
\usepackage{hhline}
\usepackage{color, colortbl}
\usepackage{multirow}
\usepackage{times}
\usepackage{epsfig}
\usepackage{lipsum} 
\usepackage{makecell}
\usepackage{url}

\newcommand{\cmark}{\ding{51}}%
\newcommand{\xmark}{\ding{55}}%

\usepackage[pagebackref,breaklinks,colorlinks]{hyperref}

\usepackage[capitalize]{cleveref}
\crefname{section}{Sec.}{Secs.}
\Crefname{section}{Section}{Sections}
\Crefname{table}{Table}{Tables}
\crefname{table}{Tab.}{Tabs.}

\def\cvprPaperID{4764} %
\def\confName{CVPR}
\def\confYear{2023}

\hypersetup{
urlcolor=blue,
}
\begin{document}

\title{Logical Consistency and Greater Descriptive Power for Facial Hair Attribute Learning}

\author{Haiyu Wu$^{1}$, Grace Bezold$^{1}$, Aman Bhatta$^{1}$, Kevin W. Bowyer$^{1}$\\
$^{1}$University of Notre Dame\\}

\maketitle

\begin{abstract}
Face attribute research has so far used only simple binary attributes for facial hair; e.g., beard / no beard.
We have created a new, more descriptive facial hair annotation scheme and applied it to create a new facial hair attribute dataset, FH37K.
Face attribute research also so far has not dealt with logical consistency 
and completeness.
For example, in prior research, an image might be classified as both having no beard and also having a goatee (a type of beard).
We show that the test accuracy of previous classification methods on facial hair attribute classification drops significantly if logical consistency of classifications is enforced.
We propose a logically consistent prediction loss, LCPLoss, to aid learning of logical consistency across attributes, and also a label compensation training strategy to eliminate the problem of no positive prediction across a set of  related attributes.
Using an attribute classifier trained on FH37K, we investigate how facial hair affects face recognition accuracy, including variation across demographics.
Results show that similarity and difference in facial hairstyle have important effects on the impostor and genuine score distributions in face recognition.
The code is at \url{https://github.com/HaiyuWu/LogicalConsistency}.
\end{abstract}

\vspace{-4mm}
\section{Introduction}
\label{sec:intro}

Facial attributes have been widely used in face matching/recognition \cite{berg2013poof, chan2017face, kumar2011describable, kumar2009attribute, manyam2011two, song2014exploiting}, face image retrieval \cite{li2015two, nguyen2018large}, re-identification~\cite{shi2015transferring, su2017multi, su2016deep}, training GANs~\cite{choi2018stargan, choi2020stargan, he2019attgan, li2021semantic} for generation of synthetic images, and other areas. As an important feature of the face, facial hairstyle does not attract enough attention as a research area. One reason is that current datasets have only simple binary attributes to describe facial hair, and this does not support deeper investigation. This paper introduces a more descriptive set of facial hair attributes, representing dimensions of the area of face covered, the length of the hair, and connectedness of beard/mustache/sideburns. We also propose a logically consistent predictions loss function, LCPLoss, and label compensation strategy to enhance the logical consistency of the predictions. We illustrate the use of this new, richer set of facial hair annotations by investigating the effect of beard area on face recognition accuracy across demographic groups. Contributions of this work include:

\begin{figure}[t]
    \centering
        \includegraphics[width=\linewidth]{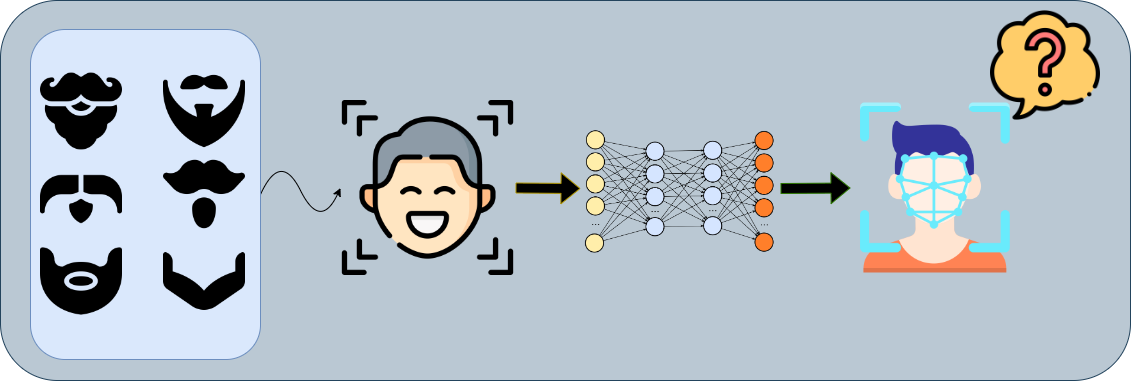}
   \caption{(1) \textit{What is the best way to define the facial hair styles?} (2) \textit{How does the facial hair classifier perform in the real-world cases?} (3) \textit{How does the face matcher treat the same (different) person with different (same) beard styles?} This paper presents our approaches and  answers for these  questions.}
\label{fig:teaser_figure}
\vspace{-2mm}
\end{figure}

\begin{itemize}
    \item A richer scheme of facial hair attributes is defined and annotations are created for the  FH37K dataset. The attributes describe facial hair features along dimensions of  area of the face covered,  length of  hair and  connectedness of elements of the hair (See Sec.~\ref{sec:fh37k} and \ref{sec:complications}).
    \item The logical consistency of classifications of the facial hair attribute classifier is analyzed. We show that the proposed LCPLoss and label compensation strategy can significantly reduce the number of logically inconsistent predictions (See Section~\ref{sec:logically_consistent_prediction} and Section~\ref{sec:facial_hair_classifier}).
    \item We analyze the effect of the beard area on face recognition accuracy. Larger difference in beard area between a pair of images matched for recognition decreases the similarity value of both impostor and genuine image pairs.
    Interestingly, the face matchers perform differently across demographic groups when image pairs have the same beard area.
    (See Section~\ref{sec:demographic_analysis})
\end{itemize}

\begin{table*}[t]
\centering
\begin{tabular}{|l|c|c|c|c|c|c|c|}
\hline
      & \# of images & \# of ids & \# of facial hair attributes   & Area & Length & CNDN & E$_{c}$ \\ \hline
\multirow{1}{*}{Berkeley Human} &
  \multirow{2}{*}{8,053} &
  \multirow{2}{*}{-} &
  \multirow{2}{*}{0} &
  \multirow{2}{*}{0} &
  \multirow{2}{*}{0} &
  \multirow{2}{*}{0} &
  \multirow{2}{*}{\xmark} \\
          Attributes\cite{berkeley_human_attributes}$^\star$                 &         &        &                                                    &   &   &   &       \\ \hline

Attributes 25K\cite{attr25k}            & 24,963  & 24,963 & 0                                 & 0 & 0 & 0 & \xmark \\ \hline
FaceTracer\cite{facetracer}$^\star$                & 15,000  & 15,000 & 1 (Mustache)                      & 0 & 0 & 0 & \xmark \\ \hline
Ego-Humans\cite{ego-humans}                & 2,714   & -      & 1 (Facial hair)                   & 0 & 0 & 0 & \xmark \\ \hline
CelebA\cite{celeba}$^\star$                    & 202,599 & 10,177 & 5 (5 o'Clock, Goatee, ...)  & 1 & 1 & 0 & \xmark \\ \hline
LFWA\cite{celeba}$^\star$                      & 13,233  & 5,749  & 5 (5 o'Clock, Goatee, ...) & 1 & 1 & 0 & \xmark \\ \hline
PubFig\cite{pubfig}$^\star$                    & 58,797  & 200    & 5 (5 o'Clock, Goatee, ...)  & 1 & 1 & 0 & \xmark \\ \hline
LFW\cite{lfw}$^\star$                       & 13,233  & 5,749  & 5 (5 o'Clock, Goatee, ...) & 1 & 1 & 0 & \xmark \\ \hline
UMD-AED\cite{umd-aed}                   & 2,800   & -    & 5 (5 o'Clock, Goatee, ...)    & 1 & 1 & 0 & \xmark \\ \hline
\multirow{1}{*}{YouTube Faces Dataset} &
  \multirow{2}{*}{3,425} &
  \multirow{2}{*}{1,595} &
  \multirow{2}{*}{5 (5 o'Clock, Goatee, ...)} &
  \multirow{2}{*}{1} &
  \multirow{2}{*}{1} &
  \multirow{2}{*}{0} &
  \multirow{2}{*}{\xmark} \\
          (with attribute labels\cite{youtube_face_dataset})                 &         &        &                                                    &   &   &   &       \\ \hline
CelebV-HQ\cite{celebv}$^\star$       & 35,666 video clips  & 15,653  & 5 (5 o'Clock, Goatee, ...) & 1 & 1 & 0 & \xmark \\ \hline
          \hline
MAAD-Face\cite{maad}$^\star$       & 3.3M  & 9,131  & 5 (5 o'Clock, Goatee, ...) & 1 & 1 & 0 & \cmark \\ \hline
\textbf{FH37K (this paper)} &  37,565 &  5,216 &  \textbf{17 (Chin area, Short...)} &  \textbf{4} &  \textbf{4} &  \textbf{4} &  \cmark \\ \hline
\end{tabular}
\caption{Comparison of facial hair descriptions in face attribute datasets.  CNDN and E$_{c}$ stand for connectedness and estimating the consistency rate of the annotations. Datasets with $\star$ are available online.  FH37K  has richer annotations that can cover the area, length, and connectedness of the facial hair.}
\label{table:statistics_of_datasets}
\vspace{-3mm}
\end{table*}

\section{Facial Hair In Face Attribute Datasets}
\label{sec:fh37k}

For a broad discussion of face attribute classification research, see the recent survey by Zheng et al~\cite{zheng2020survey}.
Here, we briefly summarize selected details of  existing facial attribute datasets, focusing on attributes describing facial hair.

Bourdev et al~\cite{berkeley_human_attributes} assembled 8,053 images from the H3D dataset~\cite{H3D} and 
the PASCAL VOC 2010 dataset~\cite{pascal_voc_2010} to create the Berkeley Human Attributes (BHA) dataset. They use
 Mechanical Turk to create 9 attributes, merging values from 5 independent annotators. Zhang et al~\cite{attr25k} collect the Attribute 25K dataset, which contains 24,963 images from 24,963 people on  Facebook. They provide 8 attributes for each image. This work, unlike most previous work in face attributes, acknowledges that some attributes may not be able to be inferred from some images.
However, how they use the "uncertain" label is not mentioned in the original paper and the dataset is not available for use. \textbf{We have an attribute called ``Info Not Vis'' and this attribute is used in our training and testing.}
Neither of \cite{berkeley_human_attributes,attr25k} includes any attribute to describe facial hair.

Kumar et al~\cite{facetracer} collected 15,000 in-the-wild face images to build the FaceTracer dataset.
The images have 10 groups of attributes including gender, age, race, environment, etc. The only attribute related to facial hair is mustache / no mustache.
Similarly, Wang et al~\cite{ego-humans} collect five million images from videos
by using the OpenCV frontal face detector to create the Ego-Humans dataset.
There are annotations for 17 face attributes, including  facial hair / no facial hair.
These two works \cite{facetracer,ego-humans} each have only a single binary attribute related to facial hair.

The Labeled Faces in the Wild\cite{lfw} (LFW) dataset has 13,233 images of cropped, aligned faces.
There are 1,680 identities in LFW that have two or more images. Kumar et al\cite{pubfig}
collected 65 attributes through Mechanical Turk~\cite{AMT} and added 8 more~\cite{kumar2011describable} for a total of 73 attributes.
Kumar et al~\cite{pubfig} also collect 58,797 images from 200 people to build the PubFig dataset. All the images
are from the internet with varied pose, lighting, expression, etc. This dataset provides
73 facial attributes. 
Liu et al~\cite{celeba} collect the largest facial attribute dataset to date, CelebA,
which has 202,599 images from 10,177 identities. It has 40 facial attributes and all the annotations are generated
by a professional labeling company. They also provide the annotations of the same attributes on the LFW dataset.
The University of Maryland Attribute Evaluation Dataset (UMD-AED) \cite{umd-aed} serves
as an evaluation dataset. It consists of 2,800 images and each attribute
has 50 positive and 50 negative samples. They use the same 40 facial attributes
as the LFWA and the CelebA datasets.
Hand et al~\cite{youtube_face_dataset} collect 3,425 frames from the original YouTube Faces Dataset. They also use the 
same 40 attributes.
Terh{\"o}rst et al~\cite{maad} created MAADFace by training a network to apply the 47 attributes across LFW and CelebA to the images from VGGFace2~\cite{vggface2}.
An interesting element of this work is that the network estimates its confidence in assigning attribute values, and about \textit{20\%} of MAADFace attribute values are left unassigned due to uncertainty.
A recent facial attributes related dataset~\cite{celebv} contains 35,666 high quality video clips. There are 83 manually labeled facial attributes covering appearance, action, and emotion.

The facial hair attributes used in existing datasets are summarized in
Table~\ref{table:statistics_of_datasets}.  The same five attributes have been used is nearly all
previous work.  A richer description of facial hairstyles is needed to
enable research into how facial hairstyle affects face recognition
accuracy. Our FH37K dataset has attributes to describe dimensions of
area of the face covered by facial hair, length of facial hair and
connectedness of parts of facial hair.  No previous work has this
level of descriptive power for facial hair, or considers the logical
consistency of the set of facial hair attributes.

\section{Overview of the FH37K Dataset}
\subsection{Dataset statistics}
FH37K contains 37,565 images, coming from a subset of CelebA\cite{celeba} and a subset of WebFace260M\cite{webface260m}. There are 5,216 identities  (3,318 identities from  CelebA and 1,898  from  WebFace260M). 
The 3,318 identities of FH37K coming from the CelebA dataset are split into train/val/test as they were in CelebA.
The identities from WebFace260M were randomly split 40\%/30\%/30\% to train/val/test.
The resulting FH37K has 28,485 images for training, 4,829 for validation, and 4,251 for testing.

All the images are manually annotated with respect to a detailed definition for each annotation, and examples and strategies for marking challenging images. The annotations of each image follow the logical relationship among the attributes.
Because subjectivity and ambiguity in assigning annotation values can only be controlled and not eliminated, we also estimate the level of consistency expected between a new annotator re-annotating the FH37K images and the annotations distributed as part of FH37K.

\subsection{Dimensions of facial hair properties}
FH37K has a larger and richer set of facial hair attributes that can be grouped into three dimensions: facial hair area, length, connectedness.
\begin{itemize}
    \item \textbf{Beard Area}: Three levels of beard area are \textit{Clean Shaven} (no beard), \textit{Chin Area} (beard limited to chin area) and \textit{Side to Side} (extending to sides of face).
    \item \textbf{Beard Length}: The five levels of length are \textit{Clean Shaven}, \textit{5 O'clock Shadow}, \textit{Short}, \textit{Medium} and \textit{Long}. The \textit{Clean Shaven} attribute can be seen as an element of description for both area and length.  %
    \item \textbf{Mustache}: Mustache-related  values are \textit{Mustache-None}, \textit{Mustache Isolated} (meaning not connected to beard) and \textit{Mustache Connected to Beard}. 
    \item \textbf{Sideburns}: Sideburns-related attribute values are \textit{Sideburns-None}, \textit{Sideburns-Present} (not connected to beard) and \textit{Sideburns Connected to Beard}.
    \item \textbf{Bald}: Bald describes scalp hair rather than facial hair, but is included in FH37K to support possible future research without needing to annotate images again. Values include \textit{Bald False}, \textit{Bald Top Only}, \textit{Bald Sides Only} and \textit{Bald Top and Sides}.
    \item \textbf{Information is not visible}: With in-the-wild imagery, it is common that information is not visible in the image to assign a value for some attribute~\cite{wu2022consistency}.
    Most previous face attribute datasets ignore this issue.  In FH37K, we use attribute values (\textit{Beard Area Info Not Vis, Beard Length Info Not Vis, Mustache Info Not Vis, Sideburns Info Not Vis, Bald Info Not Vis}).
    \vspace{-1.7mm}
\end{itemize}
More details of these 22 attributes are in Sec.~\ref{sec:complications} and the number of positive samples for each attribute is in Table 1 of the Supplementary Material. Examples of each attribute can be found in Figure 1 to 6 of Supplementary material.

\section{FH37K Data Collection}
\begin{figure*}[t]
    \centering
    \begin{subfigure}[b]{1\linewidth}
    \captionsetup[subfigure]{labelformat=empty}
        \begin{subfigure}[b]{1\linewidth}
            \begin{subfigure}[b]{0.24\linewidth}
                \begin{subfigure}[b]{0.49\linewidth}
                    \includegraphics[width=\linewidth]{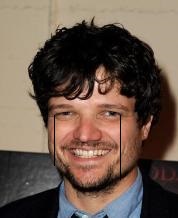}
                \end{subfigure}
                \begin{subfigure}[b]{0.49\linewidth}
                    \includegraphics[width=\linewidth]{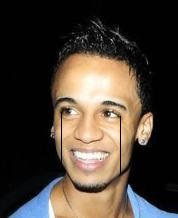}
                \end{subfigure}
            \caption{Ambiguity on Chin Area}
            \label{fig:ambiguity_on_chin_area}
            \end{subfigure}
            \begin{subfigure}[b]{0.24\linewidth}
                \begin{subfigure}[b]{0.49\linewidth}
                    \includegraphics[width=\linewidth]{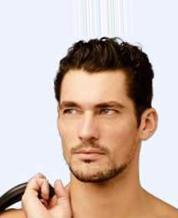}
                \end{subfigure}
                \begin{subfigure}[b]{0.49\linewidth}
                    \includegraphics[width=\linewidth]{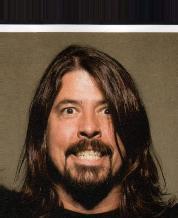}
                \end{subfigure}
            \caption{Multiple Beard Length}
            \label{fig:multiple_beard_length}
            \end{subfigure}
            \begin{subfigure}[b]{0.24\linewidth}
                \begin{subfigure}[b]{0.49\linewidth}
                    \includegraphics[width=\linewidth]{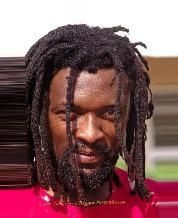}
                \end{subfigure}
                \begin{subfigure}[b]{0.49\linewidth}
                    \includegraphics[width=\linewidth]{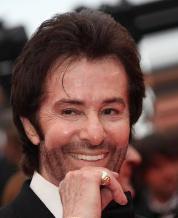}
                \end{subfigure}
            \caption{Area Info Partially Visible}
            \label{fig:area_info_partially_visible}
            \end{subfigure}
            \begin{subfigure}[b]{0.24\linewidth}
                \begin{subfigure}[b]{0.49\linewidth}
                    \includegraphics[width=\linewidth]{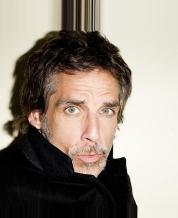}
                \end{subfigure}
                \begin{subfigure}[b]{0.49\linewidth}
                    \includegraphics[width=\linewidth]{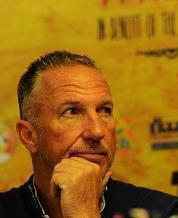}
                \end{subfigure}
            \caption{Length Info partially visible}
            \label{fig:length_info_partially_visible}
            \end{subfigure}
        \end{subfigure}
    \end{subfigure}
   \caption{Example complications for marking images consistently. More examples are in Figure 7 of Supplementary Material.}
\label{fig:complication_examples}
\vspace{-3mm}
\end{figure*}
Images in FH37K are cropped and aligned.
For CelebA, images with distributed CelebA annotations of No\_beard=false were reviewed for possible inclusion in FH37K and new annotations.
A large fraction of CelebA images with a No\_beard=false annotation actually did not have facial hair and were dropped from FH37K, and 253 images with a No\_beard=false annotation actually did not contain a face and were dropped.
CelebA images kept for FH37K were manually annotated.
Annotators read a document containing definitions and examples of the FH37K annotations before annotating and were encouraged to refer to the document as needed.
Images from CelebA had a low number of positive examples of some FH37K attributes.
A classifier was trained using this data and run on WebFace260M to generate images of additional identities, with a focus on increasing the initially under-represented positive examples.
The 4,274 images selected from WebFace260M resulted in all attributes except bald only on sides and long beard having at least 1,000 images.
The images from WebFace260M were then manually assigned attribute values in the same way as for CelebA.
The result is FH37K, 37,565 images with an aggregate total of 0.8M annotations.

\subsection{Complications for Consistent Annotation}
\label{sec:complications}
Consistent annotations that align the content of the images and the concept to be learned is an important element of any machine learning dataset.
To ensure that each annotator is oriented to the same concept for each attribute, we provided a document with detailed definition and examples for each attribute. 
However, there are still difficulties to mark annotations consistently on these in-the-wild images.

Figure~\ref{fig:complication_examples} shows four main complications: ambiguous definition of ``chin area'', varying beard length, beard area information partially visible, and beard length information partially visible. 
Without an explicit definition, ``chin area'' is subjective and can be interpreted differently by different annotators.
To address this, we gave annotators the specific definition that the chin area is within parallel vertical lines extending from the outer eye corners, as shown in Figure~\ref{fig:ambiguity_on_chin_area}. 
Images in Figure~\ref{fig:multiple_beard_length} show that the beard length can vary over the area of the beard.
To address this, annotators were asked to select the length value representing the longest length of the beard.
Head pose, occlusion, and lighting angle varies broadly in any in-the-wild dataset, giving rise to complications illustrated in Figures~\ref{fig:area_info_partially_visible} and ~\ref{fig:length_info_partially_visible}. A single attribute is not sufficient to describe these circumstances, and so we use the visible part plus the Info Not Vis attribute to describe these images.

To evaluate the consistency of our annotations, a fresh annotator independently annotated a random set of 1,000 images from FH37K.
This annotator had the same training documentation as the original annotators, but did not know the initial annotation values.
This annotator's results were compared to the FH37K annotations to estimate the level of agreement that a different annotator would have with the FH37K annotations.
The estimated consistency is 94.05\%. (Analysis is in Table 2 of Supplementary Material).

\section{Logically Consistent Prediction}

\label{sec:logically_consistent_prediction}
For facial attribute classification tasks, some papers group attributes based on position~\cite{ding2018deep, cao2018partially} or correlation~\cite{taherkhani2021tasks, han2017heterogeneous} to improve the accuracy on benchmarks. However, to our best knowledge, no previous  work considers the logical relationship between attributes on predictions. For example, in CelebA, (no-beard=true) and (goatee=true) would be logically inconsistent; so would (bald=true) and any of the hairstyles=true or hair colors=true, male=false and any of beard related attributes=true. In FH37K, where groups of facial hair attributes are defined to cover the range of possibilities, there is another logical constraint, where the model should give exactly one positive prediction across a set of related attributes. We formulate these issues into three categories and introduce a solution in this section.

\subsection{Logically Consistent Prediction Loss}
Consider a set of N 2D image $X = \{x_{1}, x_{2},..., x_{N}\}$ and their ground truth labels $Y = \{y_{1}, y_{2},...,y_{N}\}$, where $X \in \mathbb{R}^{D \times H\times W}$ as the $D$-dimension batch input and $Y \in \mathbb{R}^{D \times K}$ as the $D$-dimension batch output with $K$ predicted labels for each dimension. To train a multi-label classifier $f(X, W)$, Binary Cross Entropy Loss (BCELoss) is used:

\vspace{-2mm}
\begin{equation}
    \mathcal{L}_{BCE}=-\frac{1}{N}\sum^N_{i=1}y_i log(p(y_i)) + (1-y_i)log(1-p(y_i))
\label{eq:bceloss}
\end{equation}

The relative sparsity of positive labels in the multi-label classification tasks means that BCELoss guides the model to over-predict negative labels, which increases the accuracy on the benchmarks but reduces real-world utilty. Our approach is to force the model to consider the logical relationships - \textit{mutually exclusive}, \textit{collectively exhaustive}, \textit{dependency} - among groups of attributes:
\begin{itemize}
    \item \textit{mutually exclusive}: For some attribute groups, logical consistency requites that at most one can be positive. 
    \item \textit{dependency}: If attribute A is true, the attribute B must be true, otherwise the predictions are impossible.
    \item \textit{collectively exhaustive}: For some attribute groups, logical consistency requites exactly one must be positive.
\end{itemize}

Failure cases of these three relationships are in Algorithm 1 of Supplementary Material. Based on these three relationships, we propose the LCPLoss to force the model to make logically consistent predictions.

For \textit{mutually exclusive}, we formulate the sets $A_{ex}=\{attr_1, attr_2, ..., attr_{N}\}$ and $L_{ex}=\{l_{1}, l_{2}, ..., l_{N}\}$, where $l_{N}$ is the list of attributes that are mutually exclusive to the $attr_{N}$. 
Then, the probability of the mutually exclusive attributes happening at the same time is:
\begin{equation}
    \mathcal{P}_{ex}=\mathcal{P}(L_{ex}|A_{ex})P(A_{ex})
\end{equation}
For the \textit{dependency} relation, we formulate the set $A_d=\{attr_1, attr_2, ..., attr_{N}\}$ and $L_{d}=\{l_{1}, l_{2}, ..., l_{N}\}$, where $attr_N$ is the sufficient condition to the attributes in $l_{N}$. 
The function is:
\begin{equation}
    \mathcal{P}_{d}=\mathcal{P}(L_d|A_d)
\end{equation}
We formulate the calculation of $\mathcal{P}_{ex}$ and $\mathcal{P}_{d}$ as:
\begin{equation}
    \mathcal{P}=\frac{1}{N}\sum_{i=0}^{N}\mathcal{P}(\sum l_i > 0|attr_i==1)
\end{equation}
Where $l_i$ and $attr_i$ are from the binary predicted results after thresholding. Since $\mathcal{P}_{ex} \in [0, 1]$ and $\mathcal{P}_{d} \in [0, 1]$, in order to minimize $\mathcal{P}_{ex}$ and maximize $\mathcal{P}_{d}$, the LCPLoss is as:
\begin{equation}
    \mathcal{L}_{LCP}= ||\alpha \mathcal{P}_{ex} + \beta (1 - \mathcal{P}_{d})||^2
\label{eq:lcploss}
\end{equation}
Where $\alpha$ and $\beta$ are the coefficients to balance the ratio of $\mathcal{P}_{ex}$ and $\mathcal{P}_{d}$, we choose $\alpha=1$ and $\beta=24$. The final loss function is the combination of the BCELoss~\ref{eq:bceloss} and the LCPLoss~\ref{eq:lcploss}:
\begin{equation}
    \mathcal{L}_{total}=(1-\lambda)\mathcal{L}_{BCE} + \lambda\mathcal{L}_{LCP}
\end{equation}
Where $\lambda$ is the coefficient to adjust the weights of the loss, and $\lambda=0.5$ is our choice.

\subsection{Label Compensation}
The proposed LCPLoss is a solution to the impossible predictions, but it cannot handle the incomplete predictions. Hence, we propose the label compensation strategy which chooses the attribute that has the maximum confidence value in the incomplete portion as the positive prediction. For example, if none of the attributes that are related to beard area $[Clean\_Shaven=0.3, Chin\_Area=-2, Side\_to\_Side=0.1, Beard\_Area\_Info\_Not\_Vis=-1.5]$ has the confidence higher than the threshold value 0.5, then the attribute that has the highest confidence value among these attributes $Clean\_Shaven$ is the positive prediction. This strategy can eliminate all the incomplete predictions but increases the number of impossible predictions. In order to reduce this negative effect, we implement the label compensation strategy during both training and testing process. Code 1 and Code 2 in the Supplementary Material show the part of the training and testing code.

\begin{table}[t]
\centering
\begin{tabular}{|l|c|c|c|}
\hline
    model training                         & ACC$_{avg}$ & ACC$_{avg}^{n}$ & ACC$_{avg}^{p}$ \\ \hline \hline
        \multicolumn{4}{|l|}{Not considering logical consistency ...}\\ \hline
BCE                           & 88.82        & 93.72         & 54.97         \\ \hline
BCE$^*$               & 90.22        & 94.72         & 63.73         \\ \hline
BCE-MOON$^*$               & 88.96        & 90.67         & \textbf{81.75}         \\ \hline
BF$^*$               & 89.84        & 95.43         & 58.41         \\ \hline \hline
BCE + LCP                  & 88.90        & 95.55         & 46.13         \\ \hline
BCE + LCP$^*$               & 90.63        & 95.87         & 58.15         \\ \hline
BCE + LCP + LC              & 89.11        & 95.06         &  52.17        \\ \hline
BCE + LCP + LC$^*$ & \textbf{90.90}       & \textbf{95.98}         & 63.30         \\ \hline \hline
        \multicolumn{4}{|l|}{Considering logical consistency ...}\\

\hline
BCE                           & \textcolor{red}{45.10}        & \textcolor{red}{46.02}         & \textcolor{red}{32.62}         \\ \hline
BCE$^*$               & \textcolor{red}{53.29}        & \textcolor{red}{54.59}         & \textcolor{red}{42.40}         \\ \hline
BCE-MOON$^*$               & \textcolor{red}{46.46}        & \textcolor{red}{47.54}         & \textcolor{red}{32.95}         \\ \hline
BF$^*$               & \textcolor{red}{39.96}        & \textcolor{red}{40.95}         & \textcolor{red}{31.45}         \\ \hline \hline
BCE + LCP                           & \textcolor{red}{27.66}        & \textcolor{red}{28.19}         & \textcolor{red}{18.80}         \\ \hline
BCE + LCP$^*$               & \textcolor{red}{42.86}        & \textcolor{red}{43.70}         & \textcolor{red}{33.67}         \\ \hline \hline
\multicolumn{4}{|l|}{Label compensation on test ...}\\ \hline
BCE + LC                    & 87.47        & 90.08         & 61.55         \\ \hline
BCE + LC$^*$        & 88.83        & 91.49         & 68.78         \\ \hline
BCE-MOON + LC$^*$        & 49.39        & 50.55         & 34.62         \\ \hline
BF + LC$^*$        & 88.10        & 90.91         & 66.05         \\ \hline \hline
BCE + LCP + LC         & 87.82        & 90.37         & 59.05        \\ \hline
BCE + LCP + LC$^*$     & 89.46        & 92.02         & 66.71        \\ \hline \hline
\multicolumn{4}{|l|}{Label compensation on train and test ...}\\ \hline
BCE + LCP + LC              &  88.30       & 91.10         & 62.44         \\ \hline
BCE + LCP + LC$^*$ & \textbf{89.89}        & \textbf{92.65}         & \textbf{70.23}         \\ \hline
\end{tabular}
\caption{Accuracy of models trained with different strategies. ACC$_{avg}$ is the average accuracy for all attributes, ACC$_{avg}^{p}$ on positive samples, ACC$_{avg}^{n}$ on the negative samples. LC is the label compensation strategy. $*$ means using the transfer learning.
}
\label{table:performances_of_training_strategies}
\vspace{-2mm}
\end{table}

\begin{table}[t]
\begin{tabular}{|l|c|c|c|}
\hline
model training                            & N$_{inp}$ & N$_{imp}$ & R$_{failed}$ \\ \hline
BCE                           & 331,870                      & 1,038                          & \textcolor{red}{55.13}             \\ \hline
BCE$^*$               &  240,761                     & 6,001                        & \textcolor{red}{40.86}             \\ \hline
BCE-MOON$^*$               & 31,512                      & 313,044                        & \textcolor{red}{57.05}             \\ \hline
BF$^*$               & 339,136                      & 1,295                        & \textcolor{red}{56.37}             \\ \hline \hline
BCE + LCP                     & 470,806              & 117                          & \textcolor{red}{77.98}             \\ \hline
BCE + LCP$^*$     &  307,576                     & 300                        & \textcolor{red}{50.98}             \\ \hline \hline
\multicolumn{4}{|l|}{Label compensation on test ...}\\ \hline
BCE + LC                   & 0                            & 10,215                        & 1.69              \\ \hline
BCE + LC$^*$        & 0                            &   11,134                      &  1.84             \\ \hline  
BCE-MOON + LC$^*$        & 0                            & 330,115                        & \textcolor{red}{54.66}              \\ \hline 
BF + LC$^*$        & 0                            & 14,007                        & 2.32              \\ \hline \hline
BCE + LCP + LC             & 0                      & 14,097                          &  2.33            \\ \hline
BCE + LCP + LC$^*$       & 0                      & 6,083                        &  1.01            \\ \hline \hline
\multicolumn{4}{|l|}{Label compensation on train and test ...}\\ \hline
BCE + LCP + LC             &  0               & 7,693                        &  1.27             \\ \hline
BCE + LCP + LC$^*$ & 0                            & 5,595                        & \textbf{0.93}              \\ \hline
\end{tabular}
\caption{Results of logically consistent prediction test on a subset of WebFace260M which has 603,910 images. LC is the label compensation strategy. $*$ means using transfer learning. N$_{inp}$ is the number of the incomplete predictions. N$_{imp}$ is the number of the impossible predictions. R$_{failed}$ is the ratio of the failed cases.}
\label{table:logically_consistent_prediction}
\vspace{-2mm}
\end{table}
\section{Experiments}
In this section, we train a facial hair attribute classifier with  FH37K, and evaluate accuracy and logical consistency. 
We propose LCPLoss, and combine it with a label compensation strategy to improve the performance of logically consistent predictions on a subset of WebFace260M.
We analyze accuracy of 
ArcFace\cite{insightface, arcface} and MagFace\cite{magface} across demographics in two in-the-wild datasets.

\subsection{Facial hair attribute classifier}
\label{sec:facial_hair_classifier}

We train facial hair attribute classifiers with the ResNet50\cite{resnet} backbone, both from scratch (BCELoss and LCPLoss only) and with pretrained ImageNet \cite{imagenet} weights for transfer learning (all methods). 
We resize images to $224\times224$ and use random horizontal flip for augmentation. Batch size is 256 and the learning rate is 0.001.

We evaluate model performance both without considering the logical consistency, as traditionally done in face attribute research, and also with logical consistency. We compare the baseline BCEloss, two loss functions - Binary Focal (BF) Loss~\cite{lin2017focal}, BCE-MOON~\cite{rudd2016moon} - that handle the imbalanced dataset problem, and the proposed LCPLoss. 

Table~\ref{table:performances_of_training_strategies} shows that, \textbf{before considering the logical consistency on predictions}, BCE-MOON outperforms the other methods on predicting positive labels. The proposed method has the best overall accuracy 90.78\% on average. However, \textbf{after considering the logical consistency on predictions}, the accuracy of previous methods drops significantly, 43.26\% decrease on average. The accuracy of the proposed method decreases from 90.90\% to 89.89\% for transfer learning training strategy, and from 89.11\% to 88.30\% for training from scratch.

To further investigate the effect of LCPLoss, we use the label compensation strategy to complete those incomplete portions of the predictions of BCELoss scratch, BCELoss transfer learning, BCE-MOON, and BF, the accuracy increases to 87.47\%, 88.83\%, 49.39\%, 88.10\% respectively. It reflects that labeling images in a logically consistent way can guide the model learning to a consistent pattern on-the-fly. In addition, the methods for handling the imbalanced data could make a high-accuracy illusion without considering the logical consistency on prediction, e.g. the accuracy of BCE-MOON decreases from 81.75\% to 34.62\% on positive side. The performance of our model on each attribute is in Table 3 of the Supplementary Material.

To show the importance of logically consistent prediction of the model, we use the images of the first 30,000 identities in the sub-folder 0 from the WebFace260M dataset as a test set. Table~\ref{table:logically_consistent_prediction} shows that, on average, 52.35\% of the predictions generated by the BCE, BCE-MOON, and BF methods are logically inconsistent. After adding the label compensation strategy, the failure rates decrease dramatically. The proposed LCPLoss has the lowest fail rate 0.93\%. Note that, more incomplete predictions will reduce the number of the impossible predictions, so comparison should consider these two numbers together rather than separately.

These results show that adding LCPLoss and label compensation strategy can significantly increase the usability of the model in  real-world cases while improving accuracy. 

\begin{table}[t]
\centering
\begin{tabular}{|c|c|c|c||c|}
\hline
Demographic &
  CS/\% &
  CA/\% &
  S2S/\% &
  \multicolumn{1}{c|}{Total} \\ \hline
\multirow{2}{*}{Asian Male} &
  \begin{tabular}[c]{@{}c@{}}21,374\\ / 75.17\end{tabular} &
  \begin{tabular}[c]{@{}c@{}}6,336\\ / 22.28\end{tabular} &
  \begin{tabular}[c]{@{}c@{}}726\\ / 2.55\end{tabular} &
  \multicolumn{1}{c|}{28,436} \\ \cline{2-5} 
 &
  \begin{tabular}[c]{@{}c@{}}15,378\\ / 84.38\end{tabular} &
  \begin{tabular}[c]{@{}c@{}}1,922\\ / 10.54\end{tabular} &
  \begin{tabular}[c]{@{}c@{}}925\\ / 5.08\end{tabular} &
  18,225 \\ \hline
\multirow{2}{*}{Black Male} &
  \begin{tabular}[c]{@{}c@{}}5,529\\ / 21.63\end{tabular} &
  \begin{tabular}[c]{@{}c@{}}11,784\\ / 46.10\end{tabular} &
  \begin{tabular}[c]{@{}c@{}}8,247\\ / 32.27\end{tabular} &
  25,560 \\ \cline{2-5} 
 &
  \begin{tabular}[c]{@{}c@{}}3,539\\ / 33.64\end{tabular} &
  \begin{tabular}[c]{@{}c@{}}3,322\\ / 31.58\end{tabular} &
  \begin{tabular}[c]{@{}c@{}}3,658\\ / 34.78\end{tabular} &
  10,519 \\ \hline
\multirow{2}{*}{Indian Male} &
  \begin{tabular}[c]{@{}c@{}}12,697\\ / 51.07\end{tabular} &
  \begin{tabular}[c]{@{}c@{}}3,368\\ / 13.55\end{tabular} &
  \begin{tabular}[c]{@{}c@{}}8,795\\ / 35.38\end{tabular} &
  24,860 \\ \cline{2-5} 
 &
  \begin{tabular}[c]{@{}c@{}}7,654\\ / 48.47\end{tabular} &
  \begin{tabular}[c]{@{}c@{}}1,781\\ / 11.28\end{tabular} &
  \begin{tabular}[c]{@{}c@{}}6,356\\ / 40.25\end{tabular} &
  15,791 \\ \hline
\multirow{2}{*}{White Male} &
  \begin{tabular}[c]{@{}c@{}}13,265\\ / 54.7\end{tabular} &
  \begin{tabular}[c]{@{}c@{}}4,823\\ / 19.89\end{tabular} &
  \begin{tabular}[c]{@{}c@{}}6,162\\ / 25.41\end{tabular} &
  24,250 \\ \cline{2-5} 
 &
  \begin{tabular}[c]{@{}c@{}}25,980\\ / 63.47\end{tabular} &
  \begin{tabular}[c]{@{}c@{}}3,668\\ / 8.96\end{tabular} &
  \begin{tabular}[c]{@{}c@{}}11,287\\ / 27.57\end{tabular} &
  40,935 \\ \hline
\end{tabular}
\caption{High-confidence ($\geq 0.9$) beard area predictions for BUPT-B (top number) and BA-test (bottom number) images, broken out by prediction and demographic.}
\label{table:number_of_picked_image}
\end{table}

\subsection{Annotations and Recognition Accuracy}

\label{sec:demographic_analysis}
Experiments presented in this section show the potential value of accurate facial hair annotations in adaptive thresholding for recognition accuracy.
ArcFace and MagFace are used to extract the feature vectors. 
Previous work \cite{Abdurrahim_2018, Albiero_2022, krishnapriya2020issues, albiero2020does1, 9650887, albiero2020does, bhatta2022gender, wu2022face, terhorst2021comprehensive} shows that biases exist across gender, age, and race. In order to reduce the impact of these factors, the BUPT-Balancedface (BUPT-B)~\cite{Wang_2019_ICCV} and BA-test datasets are used. BUPT-B has 1.3M images from Asian (A), Black (B), Indian (I), White (W). Each ethnicity has 7,000 identities. Since it does not have gender information, we use FairFace~\cite{karkkainen2021fairface} to predict the gender for each identity. BA-test is a bias-aware test set we assembled on VGGFace2~\cite{vggface2}, which has 665,562 face images from 8,870 identities. It groups the people into A, B, I, W, and by gender (M,F). Images from BA-test are samples from VGGFace2~\cite{vggface2} with the head pose, image quality, brightness balanced. The gender and ethnicity labels are predicted by FairFace. The images in these two datasets are cropped and aligned by using img2pose~\cite{img2pose}.

\begin{figure*}[t]
    \centering
    \begin{subfigure}[b]{1\linewidth}
    \captionsetup[subfigure]{labelformat=empty}
        \begin{subfigure}[b]{1\linewidth}
            \begin{subfigure}[b]{1\linewidth}
                \begin{subfigure}[b]{0.24\linewidth}
                    \includegraphics[width=\linewidth]{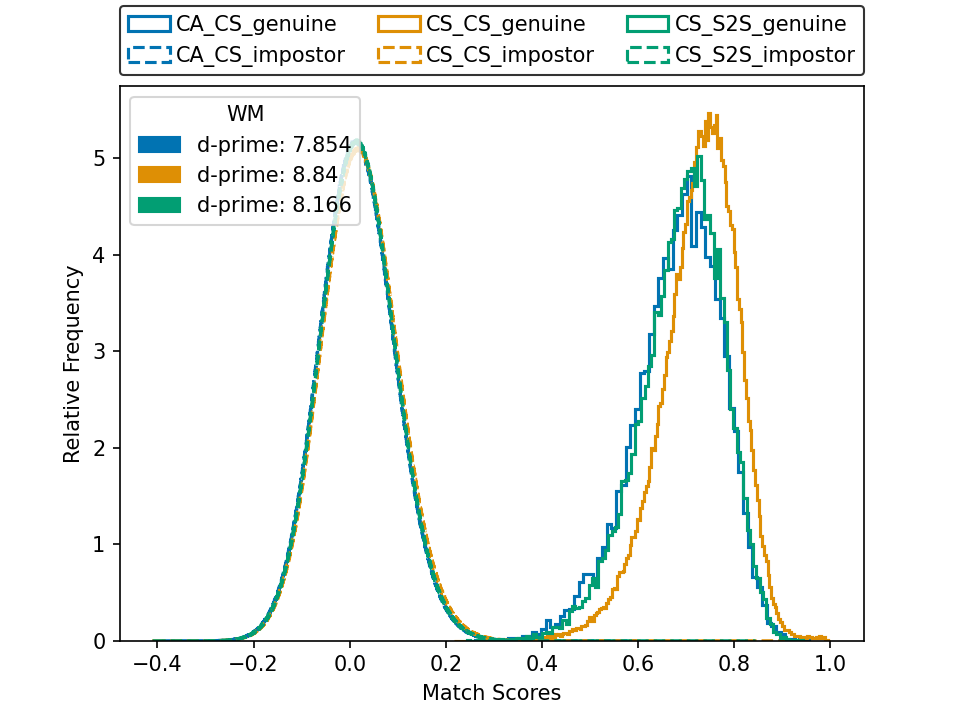}
                \end{subfigure}
                \begin{subfigure}[b]{0.24\linewidth}
                    \includegraphics[width=\linewidth]{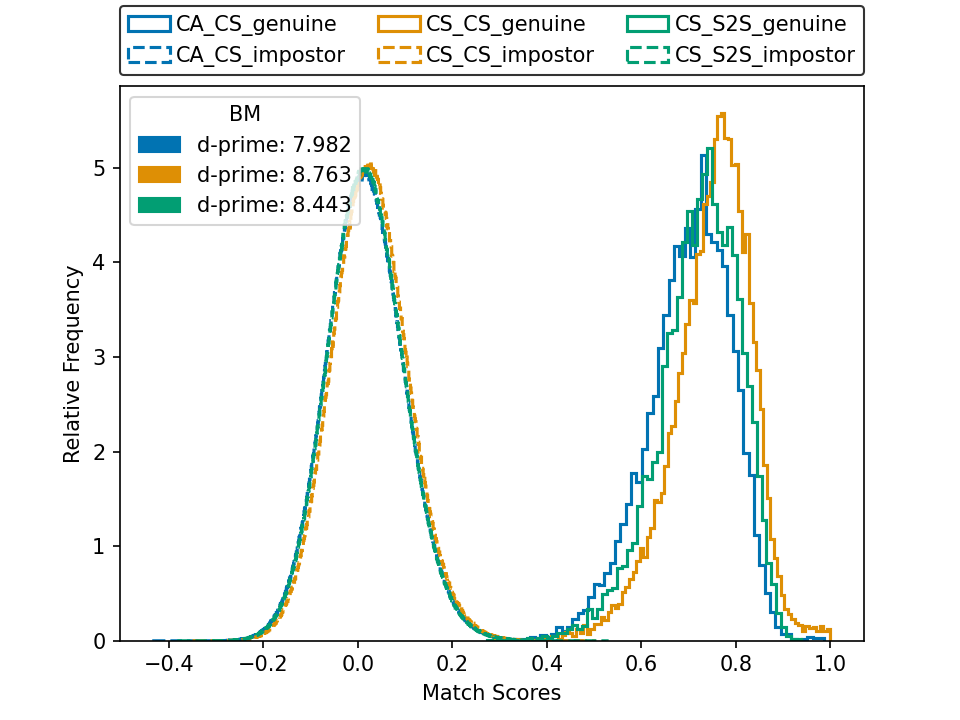}
                \end{subfigure}
                \begin{subfigure}[b]{0.24\linewidth}
                    \includegraphics[width=\linewidth]{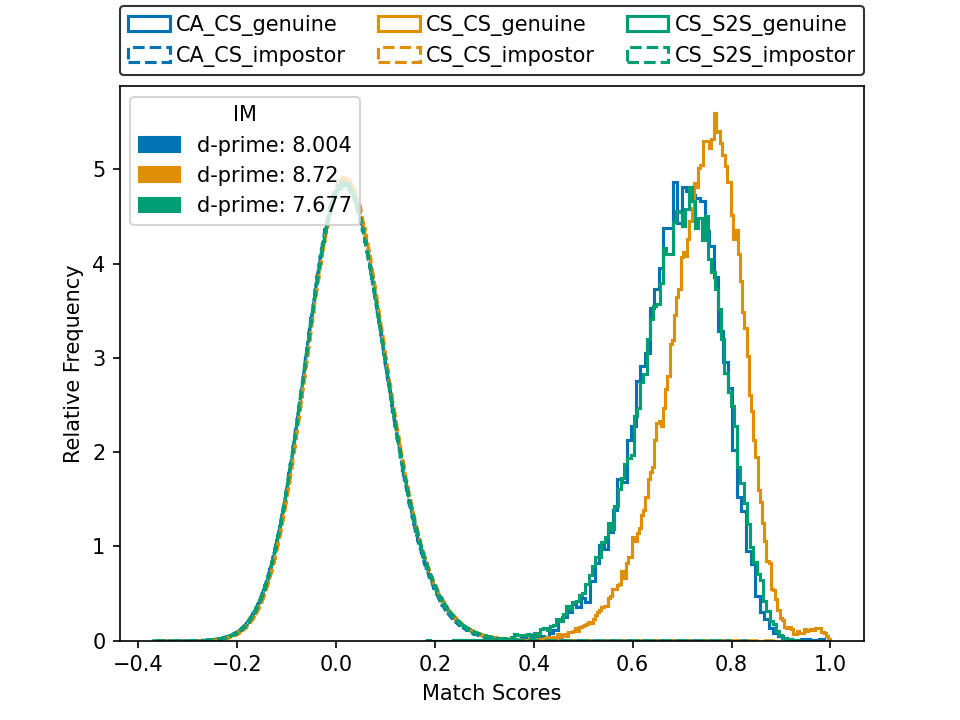}
                \end{subfigure}
                \begin{subfigure}[b]{0.24\linewidth}
                    \includegraphics[width=\linewidth]{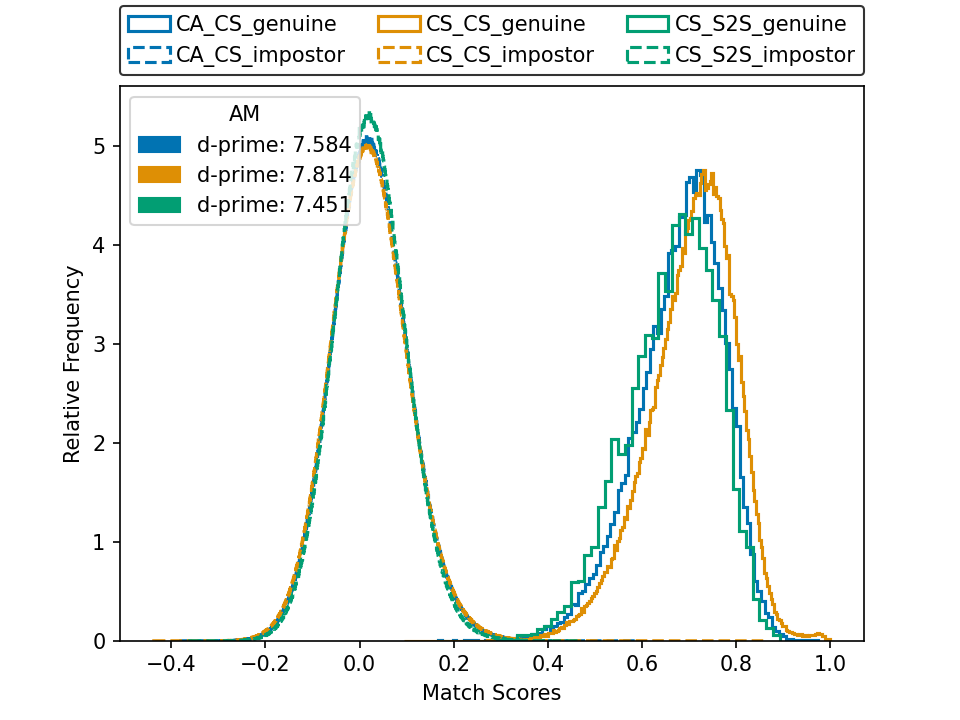}
                \end{subfigure}
            \end{subfigure}
            \begin{subfigure}[b]{1\linewidth}
                \begin{subfigure}[b]{0.24\linewidth}
                    \includegraphics[width=\linewidth]{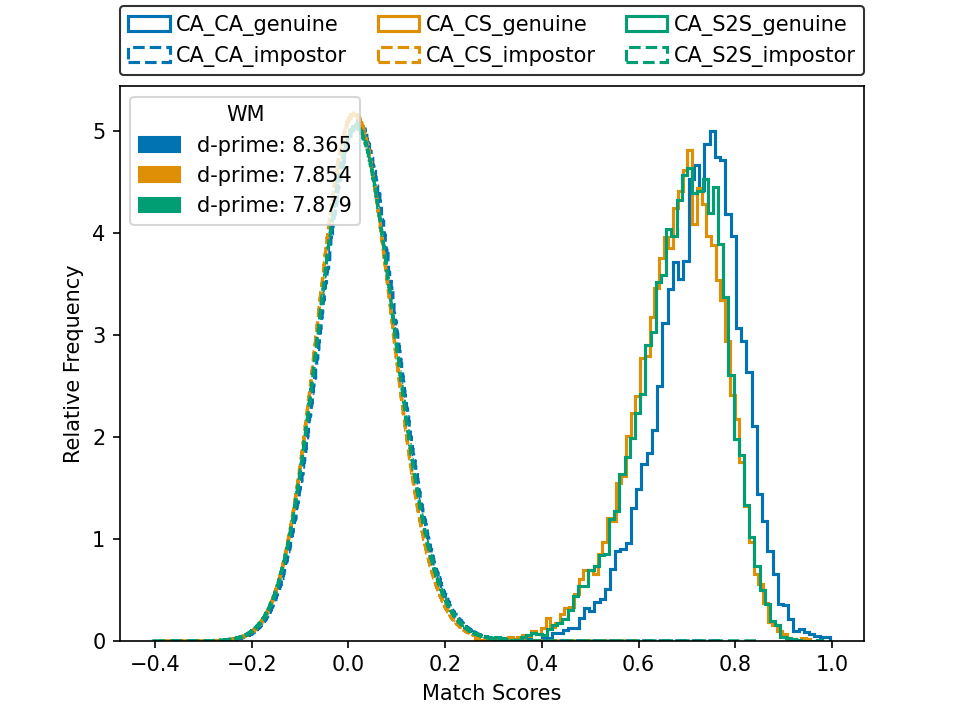}
                \end{subfigure}
                \begin{subfigure}[b]{0.24\linewidth}
                    \includegraphics[width=\linewidth]{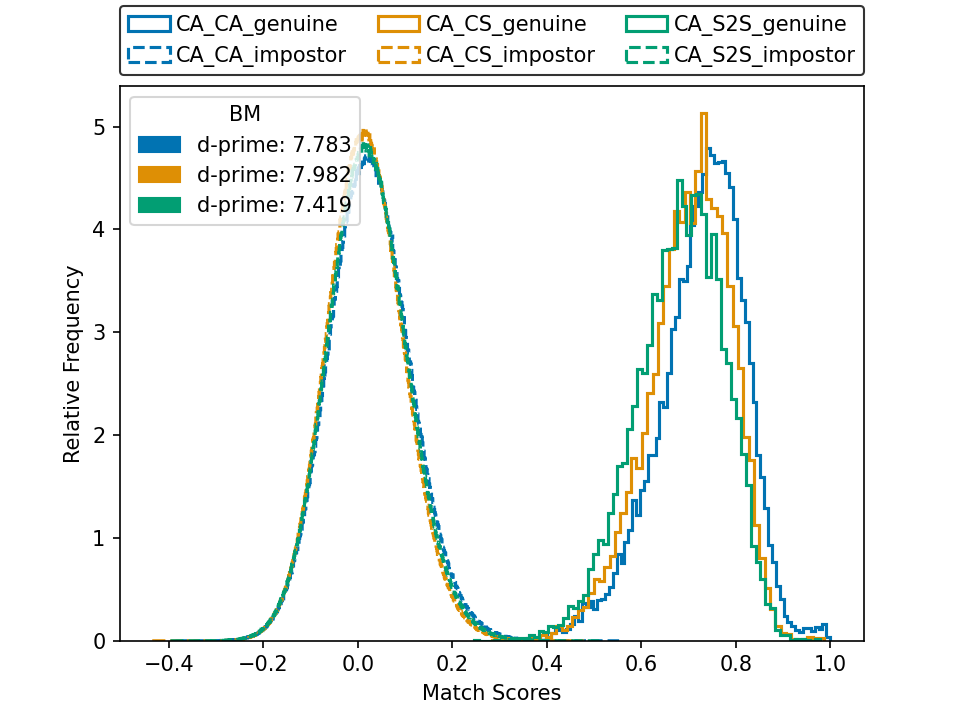}
                \end{subfigure}
                \begin{subfigure}[b]{0.24\linewidth}
                    \includegraphics[width=\linewidth]{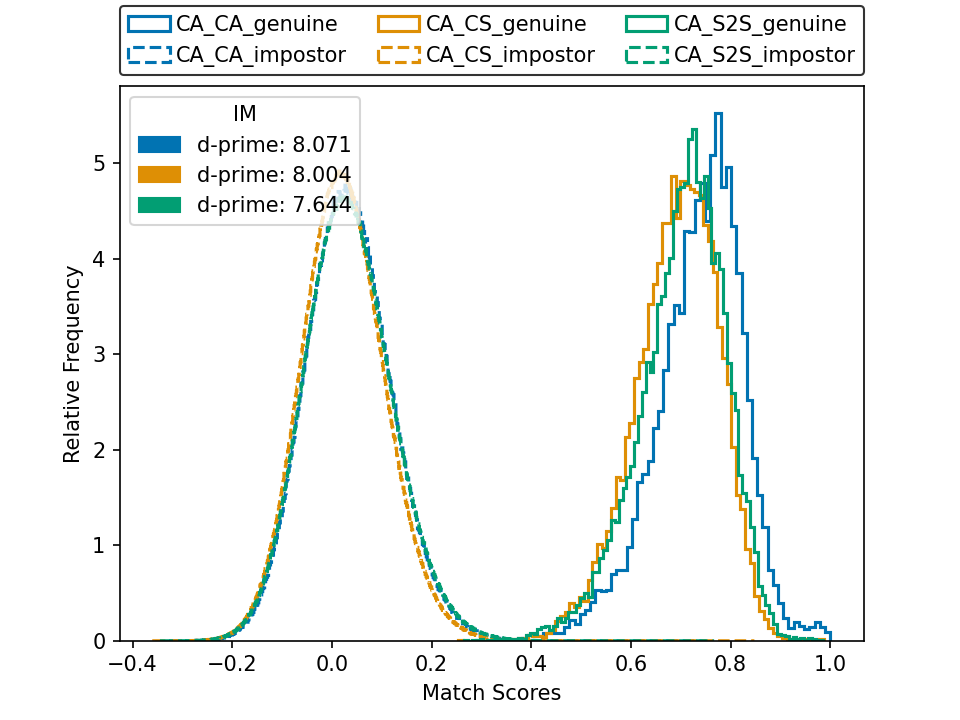}
                \end{subfigure}
                \begin{subfigure}[b]{0.24\linewidth}
                    \includegraphics[width=\linewidth]{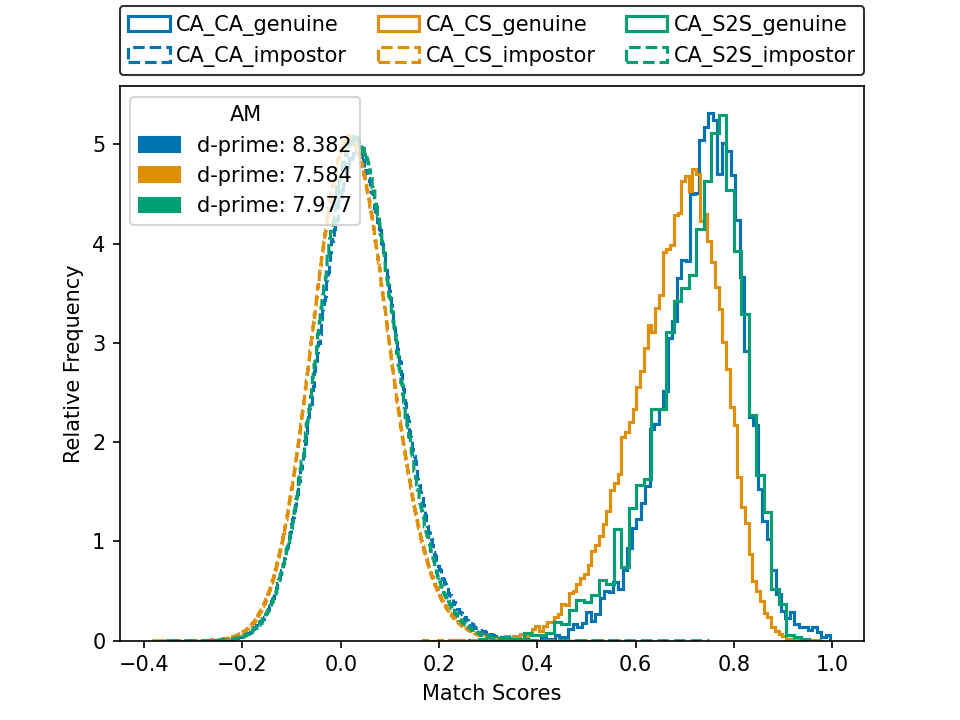}
                \end{subfigure}
            \end{subfigure}
        \end{subfigure}
        \begin{subfigure}[b]{1\linewidth}
            \begin{subfigure}[b]{1\linewidth}
                \begin{subfigure}[b]{0.24\linewidth}
                    \includegraphics[width=\linewidth]{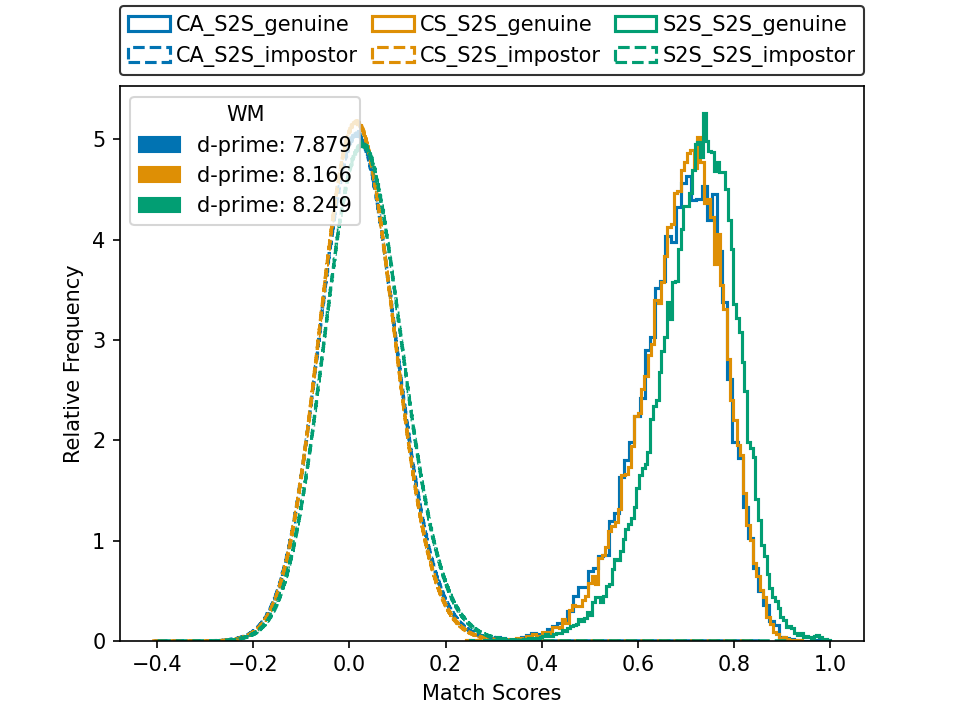}
                \end{subfigure}
                \begin{subfigure}[b]{0.24\linewidth}
                    \includegraphics[width=\linewidth]{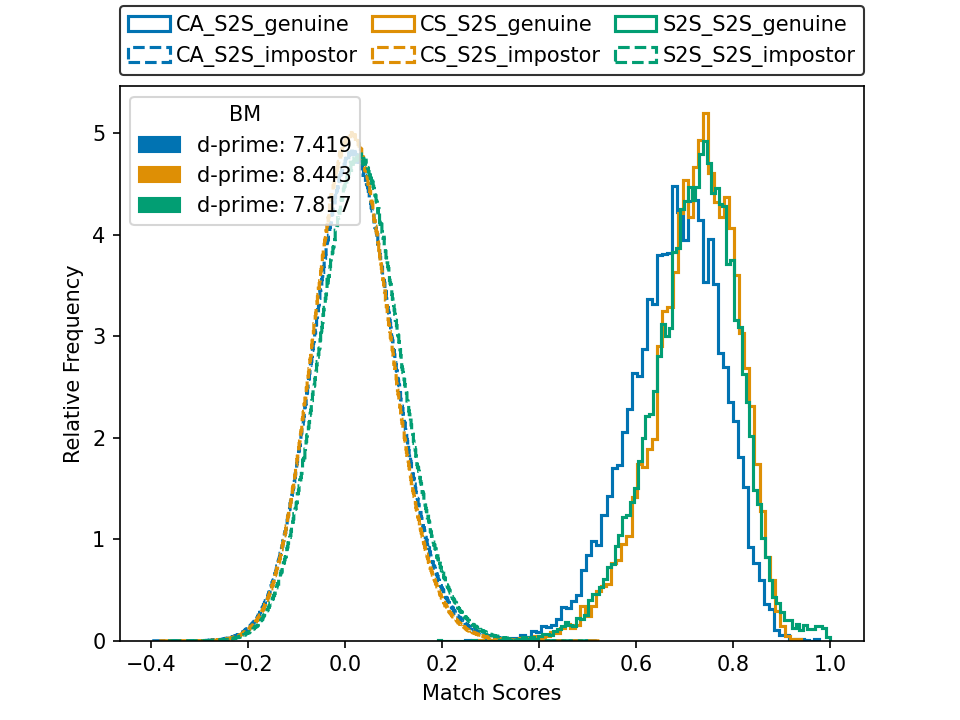}
                \end{subfigure}
                \begin{subfigure}[b]{0.24\linewidth}
                    \includegraphics[width=\linewidth]{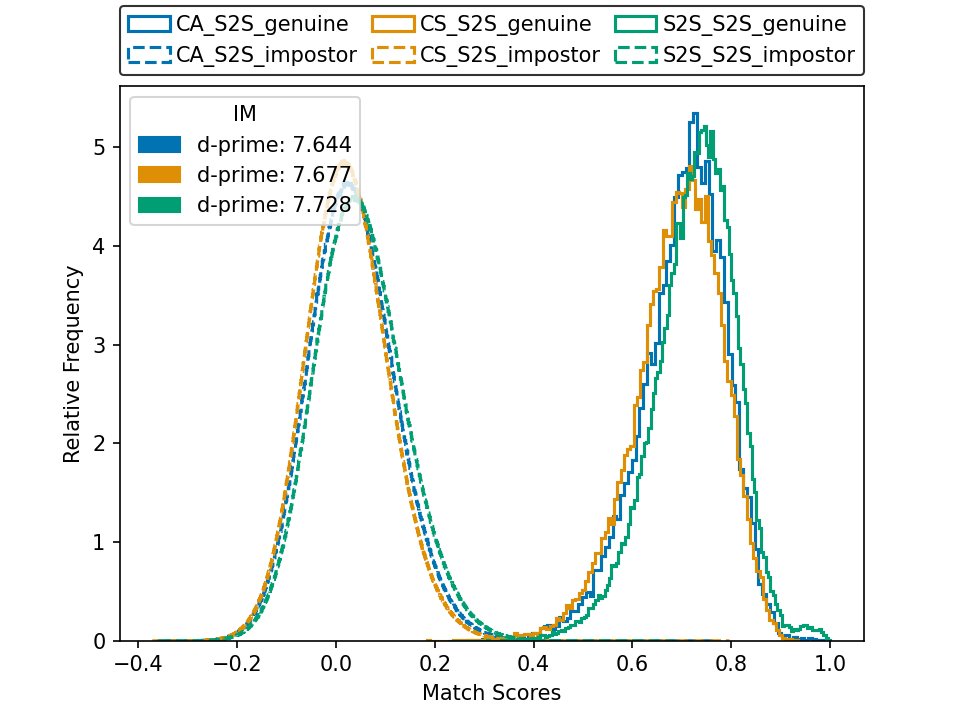}
                \end{subfigure}
                \begin{subfigure}[b]{0.24\linewidth}
                    \includegraphics[width=\linewidth]{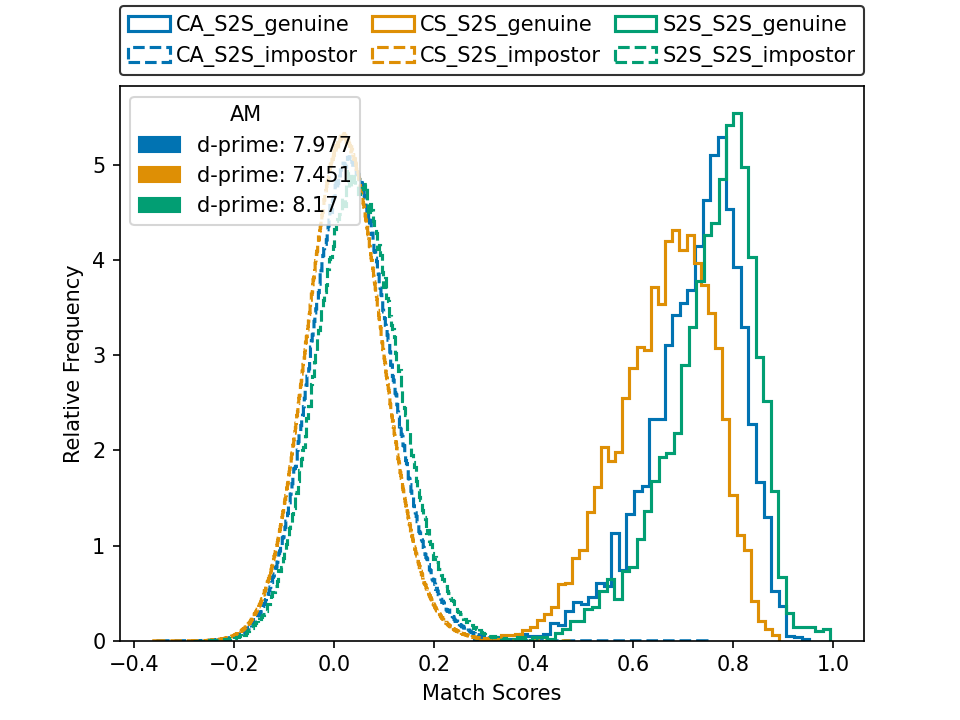}
                \end{subfigure}
            \end{subfigure}
            \begin{subfigure}[b]{1\linewidth}
                \begin{subfigure}[b]{0.49\linewidth}
                    \begin{subfigure}[b]{0.49\linewidth}
                        \includegraphics[width=\linewidth]{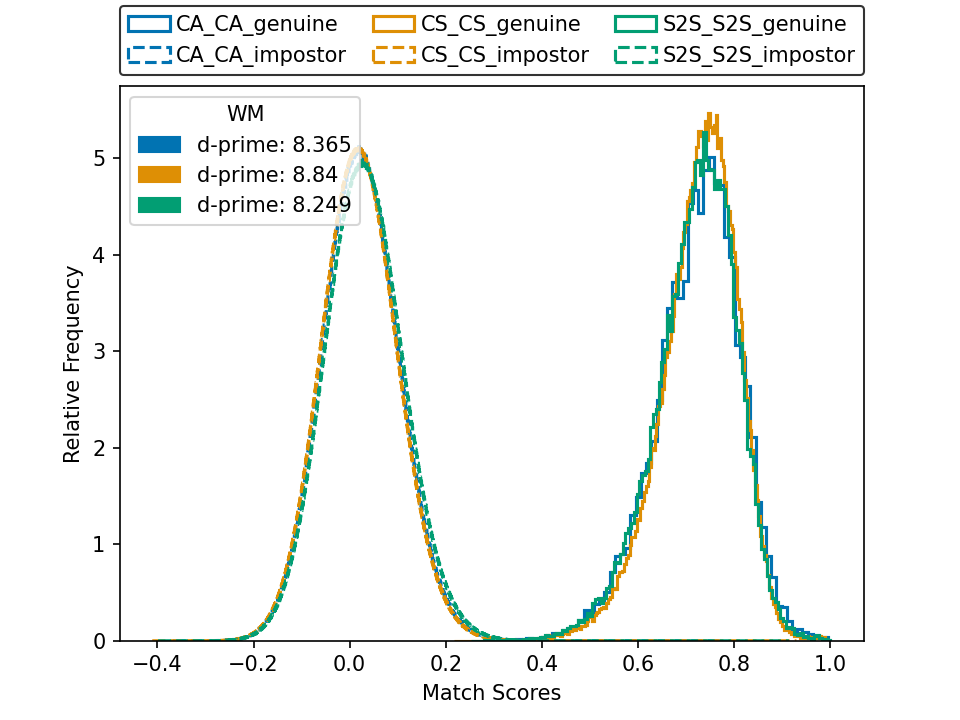}
                    \end{subfigure}
                    \begin{subfigure}[b]{0.49\linewidth}
                        \includegraphics[width=\linewidth]{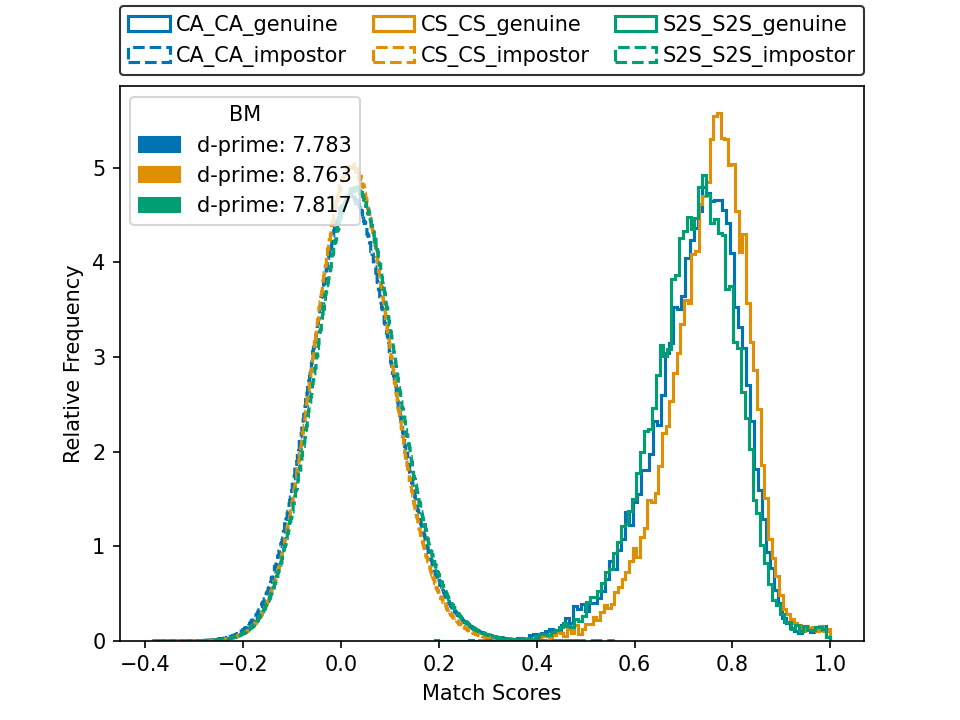}
                    \end{subfigure}
                \end{subfigure}
                \begin{subfigure}[b]{0.49\linewidth}
                    \begin{subfigure}[b]{0.49\linewidth}
                        \includegraphics[width=\linewidth]{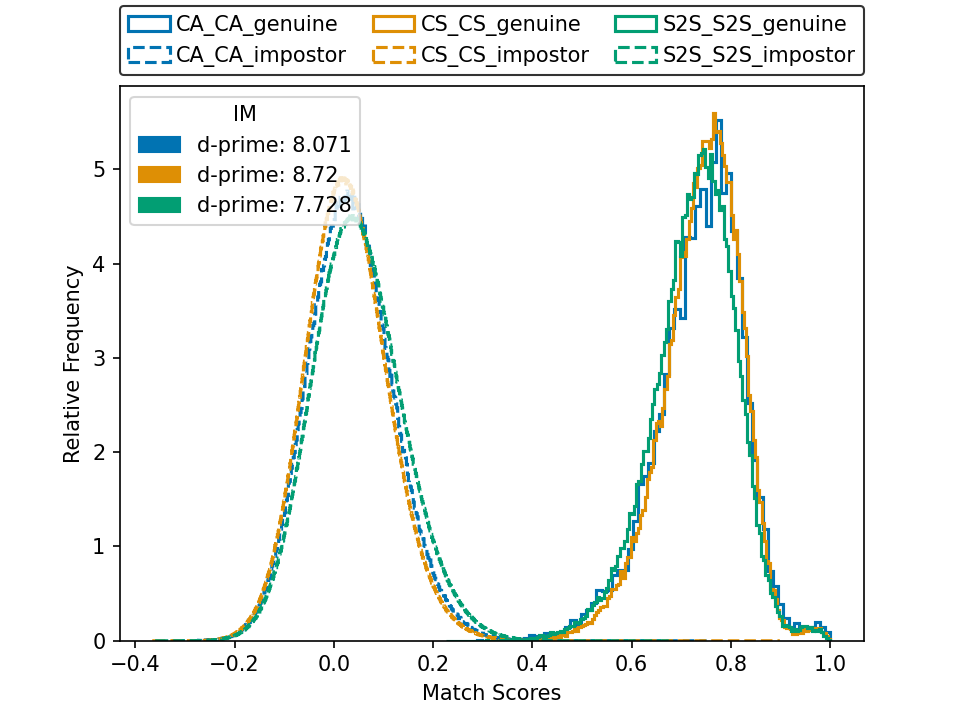}
                    \end{subfigure}
                    \begin{subfigure}[b]{0.49\linewidth}
                        \includegraphics[width=\linewidth]{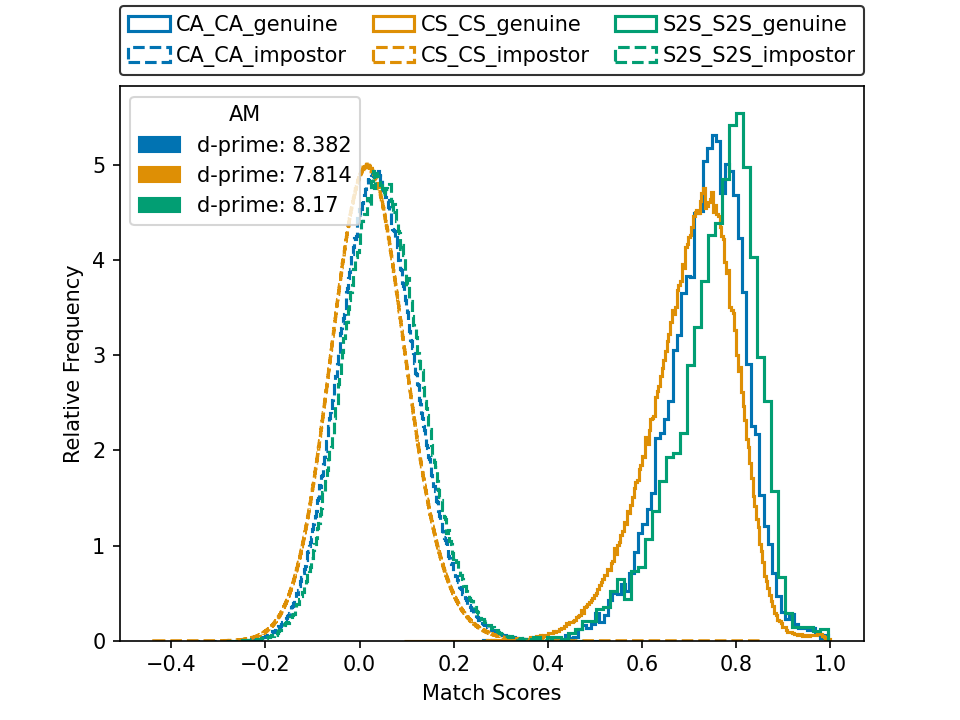}
                    \end{subfigure}
                \end{subfigure}
            \end{subfigure}
        \end{subfigure}
    \end{subfigure}
   \caption{Facial hair attribute based genuine and impostor distributions for WM, BM, IM, AM in BA-test dataset. First row is CS focused plots, second is CA focused plots,  third is S2S focused, and last row is same-beard-area focused plots. The feature extractor is MagFace.}
\label{fig:facial_hair_demographic_analysis-magface-BA-test}
\end{figure*}

For each matcher, we compute the impostor distribution separately for each demographic, and then select the threshold for a 1-in-10,000 FMR for the Caucasian male demographic as the threshold for all demographics. This follows the NIST report on demographic effects in face recognition accuracy~\cite{FRVT_2019_Part3}. Also, this method makes the cross demographic differences in FMR more readily apparent.

Facial hair is a male characteristic in general. To investigate how beard area affects accuracy across demographic groups, we first select images
with Clean Shaven (CS), Chin Area (CA), or Side to Side (S2S) beard area, using 0.9 as the threshold to pick the high-confidence samples. There are six categories of image pairs based on beard area: (CA,CA), (CA,CS), (CA,S2S), (CS,CS), (CS,S2S), and (S2S,S2S). The number of image pairs varies greatly across facial hair categories and
demographic. The number of images selected from each demographic group is in Table~\ref{table:number_of_picked_image}.

Figure~\ref{fig:facial_hair_demographic_analysis-magface-BA-test} (and Figures 1, 2, 3 of Supplementary Material) shows the impostor and genuine distributions of WM, BM, IM, and AM from BA-test and BUPT-B. 
As a general conclusion for both matchers and both datasets, beard area has more effect on the genuine distribution than the impostor. 
Image pairs with larger difference in beard area have lower similarity, and image pairs with the same beard area attribute have higher similarity.
For instance, in the CS focused plots, (CS,CS) has highest similarity and (CS,S2S) has lowest similarity. 
For the image pairs that have the same beard area, the matchers perform differently for WM, BM, IM, and IM across the datasets. (CS,CS) has highest similarity in general. However, (CA,CA) and (S2S,S2S) have a larger difference on WM and AM than BM and IM in the BUPT-B dataset for both matchers. (S2S,S2S) has highest similarity and (CS,CS) has lowest similarity for AM in BA-test dataset for both matchers.

\begin{table*}[ht]
\centering
\begin{tabular}{|c|c|c|c|c|c|c|c|c|}
\hline
\multicolumn{1}{|c|}{BA-test} &
  N$_{pairs}$ &
  AM &
  N$_{pairs}$ &
  IM &
  N$_{pairs}$ &
  BM &
  N$_{pairs}$ &
  WM \\ \hline
\multirow{2}{*}{(CA,CA)} &
  \multirow{2}{*}{1,833,490} &
  0.0558 &
  \multirow{2}{*}{1,575,614} &
  0.0567 &
  \multirow{2}{*}{5,492,657} &
  0.0571 &
  \multirow{2}{*}{6,716,047} &
  0.0142 \\ \cline{3-3} \cline{5-5} \cline{7-7} \cline{9-9} 
 &
   &
  0.0892 &
   &
  0.0849 &
   &
  0.0742 &
   &
  0.0176 \\ \hline
\multirow{2}{*}{(CA,CS)} &
  \multirow{2}{*}{29,508,374} &
  0.0307 &
  \multirow{2}{*}{13,610,147} &
  0.0368 &
  \multirow{2}{*}{11,743,810} &
  0.0238 &
  \multirow{2}{*}{95,253,307} &
  0.0073 \\ \cline{3-3} \cline{5-5} \cline{7-7} \cline{9-9} 
 &
   &
  0.0435 &
   &
  0.0467 &
   &
  0.0298 &
   &
  0.0101 \\ \hline
\multirow{2}{*}{(CA,S2S)} &
  \multirow{2}{*}{1,773,388} &
  0.0277 &
  \multirow{2}{*}{11,290,041} &
  0.0745 &
  \multirow{2}{*}{12,133,642} &
  0.0418 &
  \multirow{2}{*}{41,378,314} &
  0.0116 \\ \cline{3-3} \cline{5-5} \cline{7-7} \cline{9-9} 
 &
   &
  0.0447 &
   &
  0.1111 &
   &
  0.0557 &
   &
  0.0165 \\ \hline
\multirow{2}{*}{(CS,CS)} &
  \multirow{2}{*}{117,910,749} &
  0.043 &
  \multirow{2}{*}{29,190,869} &
  0.0603 &
  \multirow{2}{*}{6,238,358} &
  0.0309 &
  \multirow{2}{*}{337,299,241} &
  0.012 \\ \cline{3-3} \cline{5-5} \cline{7-7} \cline{9-9} 
 &
   &
  0.0544 &
   &
  0.0705 &
   &
  0.0378 &
   &
  0.0153 \\ \hline
\multirow{2}{*}{(CS,S2S)} &
  \multirow{2}{*}{14,217,616} &
  0.0143 &
  \multirow{2}{*}{48,596,378} &
  0.0434 &
  \multirow{2}{*}{12,932,207} &
  0.0224 &
  \multirow{2}{*}{293,188,100} &
  0.0071 \\ \cline{3-3} \cline{5-5} \cline{7-7} \cline{9-9} 
 &
   &
  0.0207 &
   &
  0.0609 &
   &
  0.0291 &
   &
  0.0111 \\ \hline
\multirow{2}{*}{(S2S,S2S)} &
  \multirow{2}{*}{424,161} &
  0.0691 &
  \multirow{2}{*}{20,097,880} &
  0.1455 &
  \multirow{2}{*}{6,663,160} &
  0.0549 &
  \multirow{2}{*}{63,635,960} &
  0.0219 \\ \cline{3-3} \cline{5-5} \cline{7-7} \cline{9-9} 
 &
   &
  0.0861 &
   &
  0.2103 &
   &
  0.0749 &
   &
  0.0318 \\ \hline
\end{tabular}
\caption{False match rate and corresponding fraction of each beard area comparison group in BA-test. For the false match rate and fraction of each category, top number is ArcFace model, bottom is MagFace.
}
\vspace{-2mm}
\label{table:fmr-ba_test}
\end{table*}

On the impostor side, the difference is not visually obvious in Figure~\ref{fig:facial_hair_demographic_analysis-magface-BA-test} (and Figures 1, 2, 3 of Supplementary Material), so we compare false match rate (FMR) to study the effect, shown in Table~\ref{table:fmr-ba_test} and Table 4 of Supplementary Material. 
In general, images pairs with the same beard area have the highest similarity, and image pairs in the having beard vs. clean shaven pattern have the lowest similarity. For instance, the pattern of FMR of WM is: (CA,CA) $>$ (CA,S2S) $>$ (CA,CS); (CS,CS) $>$ (CS,CA) $>$ (CS,S2S); (S2S,S2S) $>$ (CA,S2S) $>$ (CS,S2S).
It is interesting that, for AM, (CA,CS) has higher FMR than (CA,S2S) and, for IM, (CA,S2S) has higher FMR than (CA,CA) in BA-test dataset. For the impostor pairs that have the same beard area, (CA,CA) $>$ (S2S, S2S) $>$ (CS,CS) is the pattern for AM, (S2S,S2S) $>$ (CS,CS) $>$ (CA,CA) is the pattern for IM, (S2S,S2S) $>$ (CA,CA) $>$ (CS,CS) is the pattern for both BM and WM. It is interesting that beard area causes different trends across demographics. Explaining these phenomena is one of our future works.

The analyses above indicates that: (1) the fraction of each type of facial hair area varies largely across demographics; for AM, over 75\% of images are clean shaven and less than 6\% of them have side to side beard area, (2) image pairs with larger difference in beard area have lower similarity, and image pairs with the same beard area have higher similarity, (3) matchers do not all have the same relative accuracy differences across demographics, and (4) the different demographics AM, IM, and WM do not follow the same relative accuracy differences across hairstyles.
In particular, the fraction of images with facial hair varies greatly across demographics.
We speculate the number of training samples of each facial hair area attribute are unbalanced and the beard length can cause this phenomenon.

\section{Conclusions and Discussion}
We introduce a more detailed scheme of facial hair description and create a dataset, FH37K, with these annotations. FH37K 
contains a threshold number of positive examples of as many of our new attributes as possible. 
The introduction of a fundamentally better dataset for exploring facial hair attributes is one contribution of this work.

We illustrate that the classifiers trained with the baseline BCELoss and the methods that handle the imbalance data have difficulty predicting logically consistent labels. As a novel approach to logical consistency in attribute learning, we introduce LCPLoss and a label compensation strategy to cause models to learn more logically consistent predictions and enforce consistency on predictions. To our best knowledge, this is the first work investigating the logical consistency on predictions in facial attribute area.
Highlighting the issues of logical consistency across attributes and introducing an approach to solve them is another contribution of this work. Our approach is not specific to facial hair, and should be generally applicable in attribute prediction.

Using our attribute model trained on FH37K, we classify images from BUPT-B and BA-test datasets, and explore how recognition accuracy is affected by facial hair.  %
One general conclusion is that image pairs with the same beard area attribute have, on average, a higher similarity score, for both impostor pairs and genuine pairs.  (Two different persons look more alike to the face matcher when they have a similar beard area.)  Similarly, image pairs with a larger difference in the beard area attribute have a lower similarity score.  
Interestingly, the pattern of change in similarity score for image pairs that are both clean-shaven, both chin-only or both side-to-side beards shows a different trend between Asian males, Indian males, Black males, and White males.
This suggests that facial hairstyle plays a subtle causal role in the widely-commented-on demographic differences in face recognition accuracy.  
Additional factors beyond the effects of facial hair may be needed to better understand demographic accuracy differences.

Possibilities for future research include improving  both accuracy and logical consistency of predictions, extending experiments on logical consistency of predictions to other multi-label classification tasks, investigating the effects of the other attributes of the facial hair on the face recognition accuracy, and exploring the explanation of demographic differences in face recognition accuracy.

\section{Acknowledgement}

Thanks to Dr. Terrance Boult, Dr. Manuel G\"unther, Dr. Emily M Hand, Dr. Wes Robbins, and teachers from RET program - Cara Storer, Nur Islam, John Gensic, Jonathan Woodard, Jill McNabnay, and Rebekah Spencer.
This research is based upon work supported in part by the Office of the Director of National Intelligence (ODNI), Intelligence Advanced Research Projects Activity (IARPA), via \textit{2022-21102100003}. The views and conclusions contained herein are those of the authors and should not be interpreted as necessarily representing the official policies, either expressed or implied, of ODNI, IARPA, or the U.S. Government. The U.S. Government is authorized to reproduce and distribute reprints for governmental purposes notwithstanding any copyright annotation therein.

{\small
\bibliographystyle{ieee_fullname}
\bibliography{egbib}
}

\end{document}